\documentclass[letterpaper]{article}
\usepackage[noblocks]{authblk}
\usepackage{aaai}
\usepackage{times}
\usepackage{helvet}
\usepackage{courier}
\usepackage{latexsym}
\usepackage{graphicx}
\usepackage{url}
\usepackage{amsfonts}
\usepackage{amsmath}
\usepackage{array}
\usepackage{xcolor}
\usepackage{float}
\usepackage{natbib}

\newcolumntype{L}{>{\centering\arraybackslash}m{4.5cm}}

\nocopyright

\title{Experiment Segmentation in Scientific Discourse as Clause-level Structured Prediction using Recurrent Neural Networks}

\author[1]{Pradeep Dasigi}
\author[2]{Gully A.P.C. Burns}
\author[1]{Eduard Hovy}
\author[3]{Anita de Waard}
\affil[1]{Language Technologies Institute \\ Carnegie Mellon University \\ 5000 Forbes Avenue \\ Pittsburgh, PA 15213 \\ USA}
\affil[2]{Information Sciences Institute \\ Viterbi School of Engineering \\ University of Southern California \\ Marina del Rey, CA 90292 \\ USA}
\affil[3]{Elsevier Research Data Services \\ Jericho, VT 05465 \\ USA}

\date{}

\begin{document}

\maketitle

\begin{abstract}
 We propose a deep learning model for identifying structure within experiment narratives
in scientific literature. We take a sequence labeling approach to this problem, and label
clauses within experiment narratives to identify the different parts of the experiment.
Our dataset consists of paragraphs taken from open access PubMed papers labeled with rhetorical information
as a result of our
 pilot annotation. Our model is a Recurrent Neural Network (RNN) with Long Short-Term Memory (LSTM) 
 cells that labels clauses. The clause representations are computed by combining word representations
 using a novel attention mechanism that involves a separate RNN.
 We compare this model against LSTMs where the input layer has simple or no attention 
 and a feature rich CRF model. Furthermore, we describe how our work could be useful for information 
 extraction from scientific literature.
 \end{abstract}

\section{Introduction}
\label{sec:introduction}
An important part of science is communicating results. There are well established rhetorical guidelines 
\citep{michael1996craft} for scientific writing that are used across disciplines and consequently, narratives describing
evidence within a scientific investigation are expected to have a certain structure. Typically, the description begins with certain background 
information which has already been proved, 
followed by some motivating hypotheses to introduce the experiment, the methods
inferences made based on those results. Understanding this structure is important since it enables the 
higher-level construction of the general argument of the paper. The reader assembles the pieces in order to
understand what was done, why it was done, what prior knowledge it builds upon and/or refutes, and with what 
certainty the final conclusions should be accepted. Without such an overall model of the experiment, 
the reader has nothing but basic assertions.

\begin{figure}
  \begin{center}
  \includegraphics[width=2.6in]{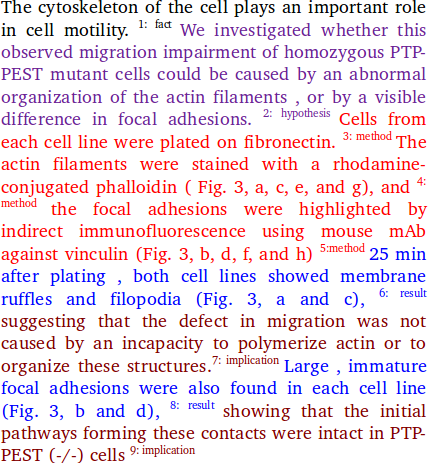}
  \caption{Example of a experiment description tagged with discourse types.}
  \label{fig:example_parse}
  \end{center}
 \end{figure}

In this work, our aim is to identify these
discourse elements given an experiment narrative. We view the task at hand as a sequence labeling problem: Given a sequence of clauses
from a paragraph describing an experiment, we seek to label the clauses with their discourse type. There exist several proposals for experiment 
discourse models \citep{liakata2010zones,nawaz2010evaluating,mizuta2004annotation,nwogu1997medical}. 
We adopt the discourse type
taxonomy 
for biological papers suggested by \cite{de2012verb}, and define our problem
as identifying the discourse type of each clause in a given experiment
description.
The taxonomy contains seven types (Table~\ref{table:taxonomy}) and
Figure~\ref{fig:example_parse} shows an example paragraph\footnote{From \textit{Angers-Loustau et al. (1999) ``Protein tyrosine phosphatase-PEST regulates focal adhesion disassembly, migration, and cytokinesis in
fibroblasts"} J. Cell Bio. 144:1019-31} broken
down into clauses and
tagged with discourse types.

While there has been some variation in the level of granularity of text in prior discourse processing work, for our task the appropriate level of processing is clearly at the clause-level. As shown in Figure~\ref{fig:example_parse}, there are many sentences in our data containing clauses of different kinds. For example, a pattern we observe frequently is when an author writes ``To understand phenomenon X, we performed experiment Y" yielding a `goal' followed by a `method' clause in a single sentence. Using the main and subordinate clauses from a Stanford parse provided good segregation of this structure. 

For this work, we focus on Systems Biology (SB) papers concerning signaling pathways in cancer cells. Typically, researchers in this field use a number of small-scale experimental assays to investigate molecular events, see \cite{Voit:2012:FCS:2341399} and \cite{svoboda_approaches_2002} for textbook and review introductions. There can easily be as many as 20-30 separate small experiments in any study that each provide evidence for the interpretive assertions being made. Our goal is to partition the text of SB papers to identify small-scale passages that describe the goals, methods, results and implications of each experiment. By convention, subfigures denoted by `1A', `1B', `1C' etc. each describe data from a separate experiment and are directly referenced in the narrative (see Figure~\ref{fig:example_parse}).

\section{Related Work}
\paragraph{Identifying structure within scientific papers}
There is a significant amount of prior work that is aimed at scientific discourse processing. \cite{teufel2002summarizing} and \cite{teufeldiscourse} describe argumentative zoning (AZ), a way of classifying scientific papers at a sentence level, into zones, thus extracting the structure from entire papers. 
\cite{hirohata2008identifying} use a 4-way classification scheme for abstracts of scientific papers for identifying objectives, methods, results and conclusions. \cite{liakata2010zones} described a three-layer finer grained annotation scheme for sentence-level annotation (with 11 separate categorical labels) for identifying the core scientific concepts of papers. Classification performance for machine learning systems to automatically tag scientific sentences was F-Score=0.51 for LibSVM classifiers \citep{liakata_automatic_2012}. There is extensive overlap between leaf elements of the CoreSC schema and our simpler discourse type model (`Hypothesis', `Goal',  `Method', and `Result' are shared between both annotation sets and tags like `Background' and `Conclusion' map to our 'fact' and 'implication' tags). 

\cite{guo2010identifying} used SVM and Na{\"i}ve Bayes classifiers to compare the three schemes described above. \cite{gupta2011analyzing} also studied the problem of extracting focus, techniques and the domain of research papers to identify the influence of research communities over each other.

In these studies, the focus of research is largely centered on modeling the discourse being used to construct a scientific argument, driving towards understanding ``sentiment expressed towards cited work, ownership of ideas, and speech acts which express rhetorical statements typical for scientific argumentation" \citep{simone_teufel_argumentative_2000}. These are driven by human-to-human communication processes of the scientific literature rather than using discourse elements to support machine reading of a semantic representation of scientific findings from primary experimental research papers. Our focus is specifically on attempting to identify text pertaining to experimental evidence for scientific IE rather focusing on authors' interpretations of those findings.

\paragraph{Deep Learning for structured prediction and text classification}
There is a great amount of work in classification and structured prediction over text and other modalities, that uses deep learning. Particularly in sequence labeling tasks in text, \citep{collobert2011natural} words are represented as vectors and used as features to train a tagger. One advantage of using this approach is the reusability of pre-trained word vectors \citep{mikolov2014word2vec,pennington2014glove} as features in various tasks. In our task, the sequences being labeled are clauses instead of words. We obtain vector representations of clauses by summarizing those of words in the clauses.

Attention has been used for complex tasks like question answering~\citep{hermann2015teaching} and machine translation \citep{bahdanau2014neural}.
In sequence-to-sequence learning problems like machine translation \citep{bahdanau2014neural}, 
parsing \citep{vinyals2015grammar} and image caption
generation \citep{xu2015show}, one network is used to encode the input modality, and a different network to decode into the output modality, with the decoder using attention to
learn parts of the input to attend to for generating a given part of the output sequence. While our work does use two different models, one for encoding clause 
representations as a function of word representations and another for decoding clause labels from clause representations, the two models
operate at different granularities.

\begin{figure}
  \begin{center}
  \includegraphics[width=3in]{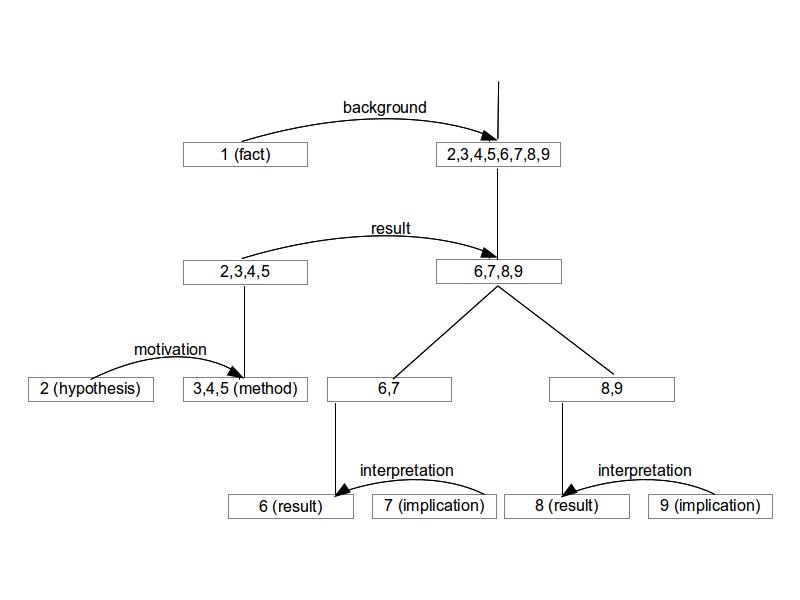}
  \caption{Tree using RST relations for the text in Figure~\ref{fig:example_parse}. The numbers indicate the clauses from the text shown in Figure~\ref{fig:example_parse}.}
  \label{fig:rst_tree}
  \end{center}
 \end{figure}

\paragraph{Comparison with RST based discourse parsing} General domain discourse parsing is a 
well-studied problem. While there are many
discourse theories (see \cite{marcu2000theory}, chapter 2 for an overview),
Rhetorical Structure
Theory (RST) by \cite{mann1988rhetorical}, received a lot of attention. It is
generally accepted that relations between non-overlapping chunks of text need to
be considered to account 
for the overall meaning \citep{marcu2000theory}. Accordingly, rhetorical
relations are central in RST for marking the structure. In contrast, the
taxonomy we use applies to the clauses 
themselves, instead of the relations between them. This is made possible by the
specificity of our domain: In the general case, it may not be possible to
identify the type of a clause 
in isolation. However, it has to be noted that the information conveyed by our
clause-centric formalism may also be expressed using a relation-centric
discourse formalism like RST. Figure~\ref{fig:rst_tree} shows one possible RST tree for the text shown in Figure~\ref{fig:example_parse}.

\begin{table}[t]
\begin{center}
  \begin{tabular}[c]{|c|L|}
 \hline
  \textbf{Type} & \textbf{Definition} \\
  \hline
  Goal & Research goal \\
  \hline
  Fact & A known fact, a statement taken to be true by the author \\
  \hline
  Result & The outcome of an experiment \\
  \hline
  Hypothesis & A claim proposed by the author \\
  \hline
  Method & Experimental method \\
  \hline
  Problem & An unresolved or contradictory issue \\
  \hline
  Implication & An interpretation of the results \\
  \hline
  \end{tabular}
\end{center}
 \caption{Seven label taxonomy from \protect\cite{de2012verb}}
 \label{table:taxonomy}
\end{table}

\section{Approach}
\begin{figure*}
  \begin{center}
  \includegraphics[width=5in]{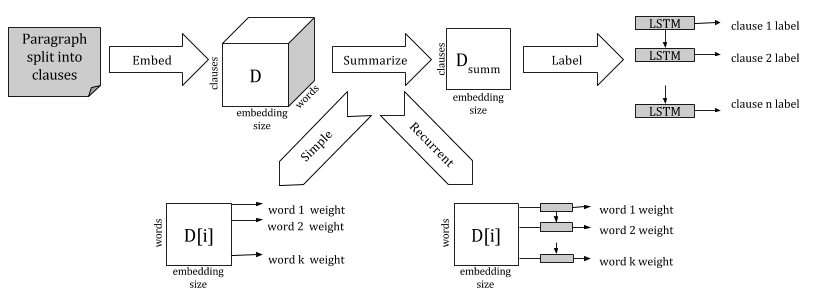}
  \caption{Our Scientific Discourse Tagging (SciDT) pipeline. The input is a list of clauses, which is \textbf{embedded} to get the tensor $D$, containing one vector per word. \textbf{Summarization} involves converting $D$ into a matrix $D_{\text{summ}}$, that has one vector per clause, which is then passed to a LSTM-RNN for \textbf{labeling}. The lower part of the figure shows the two ways of summarizing the $i^\text{th}$ clause (given by $D[i]$).}
  \label{fig:pipeline}
  \end{center}
 \end{figure*}
 We call our system Scientific Discourse Tagger (SciDT). Our pipeline is shown in Figure~\ref{fig:pipeline}. The input to the tagger is a set of clauses from a paragraph.
They are first embedded to obtain a tensor $D \in \mathbb{R}^{c \times w \times d}$ where $c$ is the number of clauses , and $w$ is the number of words in each clause, and each word is represented as
a $d$ dimensional vector.  The tensors are zero-padded along the clause and word dimensions dimensions if needed. The next step is to summarize the clause representations to obtain $D_\text{summ} \in \mathbb{R}^{c \times d}$, a matrix corresponding to the entire paragraph, with one vector per clause. Finally, $D_\text{summ}$ is fed to the Recurrent Neural Network (RNN) with Long Short-Term Memory (LSTM) \cite{hochreiter1997long} cells, to label the clauses. We propose two ways of summarizing the clause representations below. Both variants use attention to learn the weights of words within a clause based on their importance for the labeling task, and compute a weighted average of the word representations using those weights, to get the clause representation. The attention component and the LSTM-RNN are trained jointly. We use pretrained representations for words and fix them during training.
\subsection{Attention with and without context}
Both variants take as input the tensor $D$. The output in both cases is a matrix $A \in \mathbb{R}^{c \times w}$,
which contains the attention weights of all the words in the paragraph. We first project the input 
words to a lower dimensional space in both cases using a projection operator $P \in \mathbb{R}^{d \times p}$.
\begin{align}
 D_l &= tanh(D . P) &\in \mathbb{R}^{c \times w \times p}
\end{align}
The low dimension representations are then scored differently by each variant.
\paragraph{Out of context} This model defines a \textbf{simple} scoring operator $s_s \in \mathbb{R}^{p}$, that scores
each word based only on its low dimension representation. The scoring is out of context because each word is scored in isolation.
\begin{align}
		D^i_l &= D_l[i,:,:] &\in \mathbb{R}^{w \times p} \label{eq:simple_first}\\
		a^i_s &= \text{softmax}(D^i_l.s_s) &\in \mathbb{R}^{w} \label{eq:simple_att}\\
    	A_s &=
    	\begin{bmatrix}
    		a^1_s & a^2_s & \cdots & a^w_s
    	\end{bmatrix}
        &\in \mathbb{R}^{c \times w} \label{eq:simple_merge}
\end{align}
Equation~\ref{eq:simple_first} corresponds to selecting all the words in the $i^\text{th}$ clause of the paragraph. Equation~\ref{eq:simple_att} shows the computation of attention scores for all words in the clause, and equation~\ref{eq:simple_merge} simply puts the clause scores together to get the paragraph level attention values.

\paragraph{Clause context} In this variant, we score words in a clause in the context of other words that occur in the clause. Concretely, as shown in the equations below, this is a \textbf{recurrent} scoring mechanism that uses a RNN to score each word in a clause as a function of its low-dimension representation and its previous context in the clause given by the hidden layer in the RNN. It has to be noted that the recurrence in this scoring model is over words in a clause while that in the LSTM described previously is over clauses. 
\begin{align}
 D^i_l &= D_l[i,:,:] &\in \mathbb{R}^{w \times p} \label{eq:rec_first}\\
 h^i_j &= \text{tanh}(D^i_l[j,:].W_{IH} + h^i_{j-1}.W_{HH}) &\in \mathbb{R}^{w \times p} \label{eq:rec_rnn}\\
 a^i_r &= \text{softmax}(h^i_j.s_r) &\in \mathbb{R}^{w} \label{eq:rec_att}\\
 A_r &=
 \begin{bmatrix}
  a^1_r & a^2_r & \cdots & a^w_r \\
 \end{bmatrix}
 &\in \mathbb{R}^{c \times w} \label{eq:rec_merge}
\end{align}
Equation~\ref{eq:rec_first}, equation~\ref{eq:rec_att} and equation~\ref{eq:rec_merge} are similar to equation~\ref{eq:simple_first}, equation~\ref{eq:simple_att} and equation~\ref{eq:simple_merge} respectively. The operator $s_r \in \mathbb{R}^p$ is similar to $s_s$ from simple attention. In equation~\ref{eq:rec_rnn}, we apply the standard RNN recurrence to update the hidden state, using the parameters $W_{IH} \in \mathbb{R}^{p \times p}$, operating on the input word at the current timestep $j$, and $W_{HH} \in \mathbb{R}^{p \times p}$, operating on the hidden state from the previous timestep $j-1$.   
\subsection{Input to LSTM}
A weighted sum of the input tensor is computed, with the weights coming from the attention model, and it is fed to the LSTM.
\begin{align}
 D_{\text{summ}}[i,:] &= A[i,:] . D[i,:,:] &\in \mathbb{R}^{c \times d}
\end{align}
The above equation shows the composed representation of the $i$th clause stored as the $i$th row in $D_{\text{summ}}$.
\section{Experiments}
\subsection{Implementation Details}
We used the 200 dimension vectors trained on Pubmed Central data by \cite{pyysalo2013distributional} as input representations
and projected them down to 50d to keep the parameter space of the entire model under control. The projection operator is also trained along with the rest of the pipeline. LSTMs were implemented \footnote{Code is publicly available at \url{https://github.com/edvisees/sciDT}}
using Keras \citep{chollet2015keras} and attention using Theano \citep{bergstra2010theano}. We trained for atmost 100 epochs, while monitoring accuracy on held-out validation data for early stopping.  We used ADAM \citep{kingma2014adam} 
as the optimization algorithm. Dropout with $p=0.5$ was used on input to the attention layer.
\subsection{Data Preprocessing, Annotations and Pipelines}
We created a scientific discourse marked dataset from 75 papers in the area of intercellular cancer pathways taken from the Open Access\footnote{\url{http://www.ncbi.nlm.nih.gov/pmc/tools/openftlist/}} subset of Pubmed Central. 
Using a multithreaded preprocessing pipeline, we extracted the Results sections of each of those papers, and parsed all the sentences using the Stanford Parser \citep{socher2013parsing}. This process separated the main and subordinate clauses of each sentence that we process as a sequence over separate paragraphs. We asked domain experts to label each of those clauses using the seven label taxonomy suggested by \cite{de2012verb}. We also added a \textit{None} label for those clauses that do not fall under any category. Each sequence in the dataset corresponds to the clauses extracted from a paragraph. So we make the assumption that paragraphs are minimal experiment narratives. On the whole, our dataset\footnote{Please contact the authors if you would like to use this dataset for your research.} consists of 392 experiment descriptions with a total of 4497 clauses. This is an ongoing annotation effort, and we intend to make a bigger dataset in the future. 
\begin{figure}
  \begin{center}
  \includegraphics[width=3in]{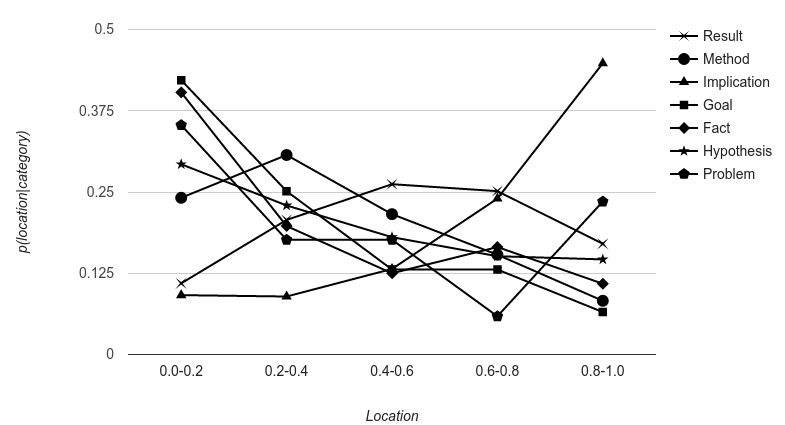}
  \caption{Chart showing probabilities of each dicourse type at various positions in a paragraph. Each paragraph is broken into five parts}
  \label{fig:loc_probs}
  \end{center}
 \end{figure}
 
Figure~\ref{fig:loc_probs} shows $p(position|type)$ values of discourse types 
at various positions in a paragraph, estimated from the entire annotated dataset. It can be 
seen that \textit{goal}, \textit{fact}, \textit{problem} and \textit{hypothesis} are more
likely at the beginning of a paragraph compared to other locations, whereas \textit{method}
peaks before the middle, \textit{result} at the middle, and \textit{implication} clearly towards
the end of a paragraph. This trend supports the expected narrative structure described in 
Section 1.

\begin{table}
\begin{center}
  \begin{tabular}[c]{|c|c|c|c|}
 \hline
  Model & Attention & Accuracy & F-score \\
  \hline
  CRF & - & 0.6942 & 0.6818 \\
  SciDT & None & 0.6912 & 0.6749\\
  \hline
  SciDT & Simple & 0.7379 & 0.7261 \\
  SciDT & Recurrent & \textbf{0.7502} & \textbf{0.7405}\\
  \hline
  \end{tabular}
\end{center}
 \caption{Accuracies and F-scores from 5-fold cross validation of SciDT in various settings and a CRF. Simple attention corresponds to the out of context variant, and recurrent to the clause context variant.}
 \label{table:results}
\end{table}
\subsection{Results and Analysis}
A baseline model we compare against is a Conditional Random Field \citep{lafferty2001conditional} that
uses as features part of speech tags, the identities of verbs and adverbs, presence of figure references and citations,
and hand-crafted lexicon features\footnote{For example, words like \textit{demonstrate} and \textit{suggest} indicate 
\textit{implication}; phrases like \textit{data not shown} indicate \textit{result}} that indicate specific discourse types.
In addition, we also test a variant of our model that does not use attention in the input layer and the clause 
vectors are obtained simply as an average of the vectors of words in them. Table~\ref{table:results} shows accuracy scores and weighted 
averages\footnote{F-scores are of all classes were averaged weighted by the number of points within that class in the gold 
standard.} of f-scores from 5-fold cross validation of the two baseline models and the two attention based models. The performance
of the averaged input SciDT is comparable to the CRF model, whereas the two attention models perform better. The performance of the 
recurrent attention SciDT model shows the importance of modeling context in attention.
 \begin{figure}[t!]
  \begin{center}
  \includegraphics[width=3.4in]{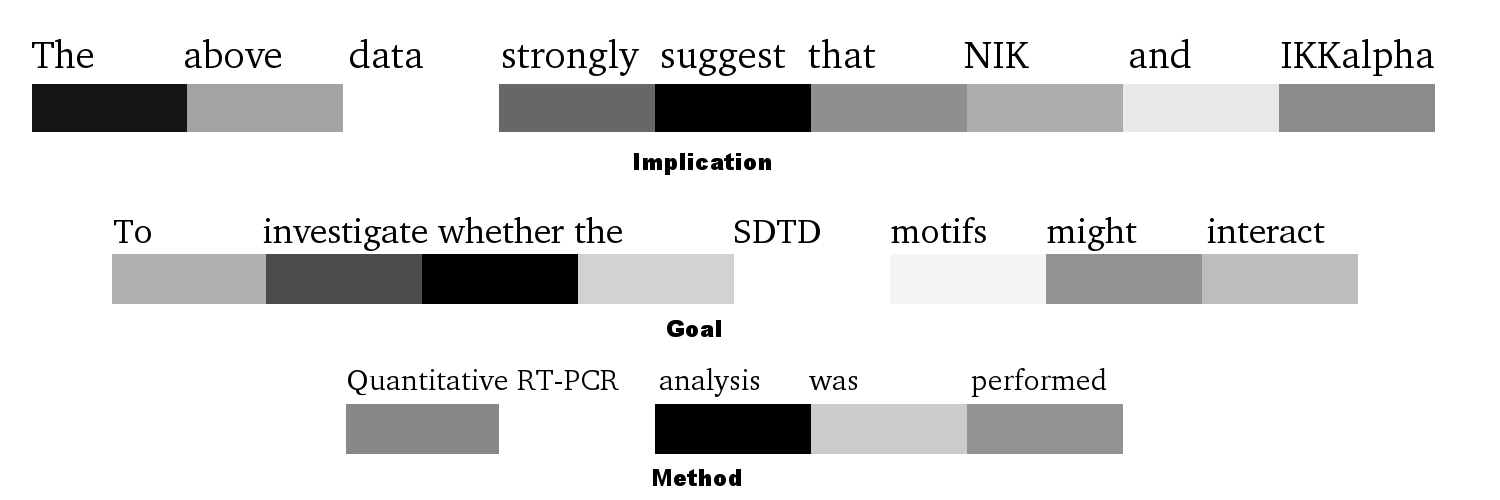}
  \caption{Examples of attention probabilities assigned by the recurrent model to words in parts of clauses and their correctly predicted labels. Darker shades show higher attention weights.}
  \label{fig:att_examples}
  \end{center}
 \end{figure}
On closer examination of the attention weights
assigned to words in unseen paragraphs, we noticed stronger trends in the recurrent attention based model. Particularly, the main verbs
of the sentences received the highest attention in many cases in the recurrent model. That the verb form is an important indicator 
of discourse type identification was shown by \cite{de2012verb}. 

Figure~\ref{fig:att_examples} shows examples of parts of clauses
and the attention weights assigned to the words. These indicate the general trends of words relevant to discourse classes being given higher attention: \textit{investigate whether} indicates
\textit{Goal}, \textit{analysis} is a \textit{Method} word, and \textit{strongly suggest} is a phrase expected in \textit{Implication}.
While some of these errors made by the LSTM may be attributed to the model itself, there were also some exceptions to the assumption that
 clauses are the smallest units of discourse. There are some infrequent cases where clauses had components of multiple discourse types in them. Moreover, the syntactic parser we used 
 to separate clauses was sometimes inaccurate, resulting in incorrect clause splits.

\section{Discussion and Conclusion}
We introduced a sequence labeling approach for identifying the discourse elements within experiment descriptions. Our model uses an attention mechanism over word representations to obtain clause representations. The results show that our attention based composition mechanism used to encode clauses adds value to the LSTM model. Visualizations show that the clause context model does indeed learn to attend to words important for the final tagging decision. In the future, we shall extend the idea of contextual attention to attend to words based on context at the paragraph level.

Our system can complement existing IE tools that operate on scientific literature, and provide useful epistemic and contextual features.  Identifying the structure of experiments provides
additional context information that can help various downstream tasks. For event
co-reference, 
one can use the structure to accurately resolve anaphora links. For example, a
reference made in an \textit{implication} statement is likely to some entity in
a \textit{result} 
statement it follows. It has to be noted that the taxonomy we used also provides
epistemic information which is helpful in information extraction (IE): IE
systems need not process
clauses labeled as \textit{hypothesis} or \textit{goal} since they do not
contain events that actually occurred. Going forward, our goal is to read, assemble and model mechanisms describing complex phenomena from collections of relevant scientific documents.

The application of our methods can reveal a small-scale discourse structure to contextualize, report and interpret evidence from individual experiments in a fine-grained context. This could be used to support cyclic models of scientific reasoning where data from individual experiments can be placed in an appropriate interpretive context within an informatics system that synthesizes knowledge across many papers. Beyond the scope of direct applications to IE, this work may be applied to Semantic Web representations of scientific knowledge and biocuration pipelines to accelerate knowledge acquisition. 

\bibliography{exp_parsing}

\begin{thebibliography}{}

\bibitem[\protect\citeauthoryear{Alley}{1996}]{michael1996craft}
Alley, M.
\newblock 1996.
\newblock {\em The craft of scientific writing}.
\newblock Springer Science \& Business Media.

\bibitem[\protect\citeauthoryear{Bahdanau, Cho, and
  Bengio}{2014}]{bahdanau2014neural}
Bahdanau, D.; Cho, K.; and Bengio, Y.
\newblock 2014.
\newblock Neural machine translation by jointly learning to align and
  translate.
\newblock {\em arXiv preprint arXiv:1409.0473}.

\bibitem[\protect\citeauthoryear{Bergstra \bgroup et al\mbox.\egroup
  }{2010}]{bergstra2010theano}
Bergstra, J.; Breuleux, O.; Bastien, F.; Lamblin, P.; Pascanu, R.; Desjardins,
  G.; Turian, J.; Warde-Farley, D.; and Bengio, Y.
\newblock 2010.
\newblock Theano: A cpu and gpu math compiler in python.
\newblock In van~der Walt, S., and Millman, J., eds., {\em Proceedings of the
  9th Python in Science Conference},  3 -- 10.

\bibitem[\protect\citeauthoryear{Chollet}{2015}]{chollet2015keras}
Chollet, F.
\newblock 2015.
\newblock Keras.
\newblock \url{https://github.com/fchollet/keras}.

\bibitem[\protect\citeauthoryear{Collobert \bgroup et al\mbox.\egroup
  }{2011}]{collobert2011natural}
Collobert, R.; Weston, J.; Bottou, L.; Karlen, M.; Kavukcuoglu, K.; and Kuksa,
  P.
\newblock 2011.
\newblock Natural language processing (almost) from scratch.
\newblock {\em Journal of Machine Learning Research} 12(Aug):2493--2537.

\bibitem[\protect\citeauthoryear{De~Waard and Pander~Maat}{2012}]{de2012verb}
De~Waard, A., and Pander~Maat, H.
\newblock 2012.
\newblock Verb form indicates discourse segment type in biological research
  papers: Experimental evidence.
\newblock {\em Journal of English for academic purposes} 11(4):357--366.

\bibitem[\protect\citeauthoryear{Guo \bgroup et al\mbox.\egroup
  }{2010}]{guo2010identifying}
Guo, Y.; Korhonen, A.; Liakata, M.; Karolinska, I.~S.; Sun, L.; and Stenius, U.
\newblock 2010.
\newblock Identifying the information structure of scientific abstracts: an
  investigation of three different schemes.
\newblock In {\em Proceedings of the 2010 Workshop on Biomedical Natural
  Language Processing},  99--107.
\newblock Association for Computational Linguistics.

\bibitem[\protect\citeauthoryear{Gupta and Manning}{2011}]{gupta2011analyzing}
Gupta, S., and Manning, C.
\newblock 2011.
\newblock Analyzing the dynamics of research by extracting key aspects of
  scientific papers.
\newblock In {\em Proceedings of 5th International Joint Conference on Natural
  Language Processing},  1--9.
\newblock Chiang Mai, Thailand: Asian Federation of Natural Language
  Processing.

\bibitem[\protect\citeauthoryear{Hermann \bgroup et al\mbox.\egroup
  }{2015}]{hermann2015teaching}
Hermann, K.~M.; Kocisky, T.; Grefenstette, E.; Espeholt, L.; Kay, W.; Suleyman,
  M.; and Blunsom, P.
\newblock 2015.
\newblock Teaching machines to read and comprehend.
\newblock In {\em Advances in Neural Information Processing Systems},
  1684--1692.

\bibitem[\protect\citeauthoryear{Hirohata \bgroup et al\mbox.\egroup
  }{2008}]{hirohata2008identifying}
Hirohata, K.; Okazaki, N.; Ananiadou, S.; Ishizuka, M.; and Biocentre, M.~I.
\newblock 2008.
\newblock Identifying sections in scientific abstracts using conditional random
  fields.

\bibitem[\protect\citeauthoryear{Hochreiter and
  Schmidhuber}{1997}]{hochreiter1997long}
Hochreiter, S., and Schmidhuber, J.
\newblock 1997.
\newblock Long short-term memory.
\newblock {\em Neural computation} 9(8):1735--1780.

\bibitem[\protect\citeauthoryear{Kingma and Ba}{2014}]{kingma2014adam}
Kingma, D., and Ba, J.
\newblock 2014.
\newblock Adam: A method for stochastic optimization.
\newblock {\em arXiv preprint arXiv:1412.6980}.

\bibitem[\protect\citeauthoryear{Lafferty, McCallum, and
  Pereira}{2001}]{lafferty2001conditional}
Lafferty, J.; McCallum, A.; and Pereira, F.~C.
\newblock 2001.
\newblock Conditional random fields: Probabilistic models for segmenting and
  labeling sequence data.

\bibitem[\protect\citeauthoryear{Liakata \bgroup et al\mbox.\egroup
  }{2012}]{liakata_automatic_2012}
Liakata, M.; Saha, S.; Dobnik, S.; Batchelor, C.; and Rebholz-Schuhmann, D.
\newblock 2012.
\newblock Automatic recognition of conceptualization zones in scientific
  articles and two life science applications.
\newblock {\em Bioinformatics (Oxford, England)} 28(7):991--1000.

\bibitem[\protect\citeauthoryear{Liakata}{2010}]{liakata2010zones}
Liakata, M.
\newblock 2010.
\newblock Zones of conceptualisation in scientific papers: a window to negative
  and speculative statements.
\newblock In {\em Proceedings of the Workshop on Negation and Speculation in
  Natural Language Processing},  1--4.
\newblock Association for Computational Linguistics.

\bibitem[\protect\citeauthoryear{Mann and Thompson}{1988}]{mann1988rhetorical}
Mann, W.~C., and Thompson, S.~A.
\newblock 1988.
\newblock Rhetorical structure theory: Toward a functional theory of text
  organization.
\newblock {\em Text-Interdisciplinary Journal for the Study of Discourse}
  8(3):243--281.

\bibitem[\protect\citeauthoryear{Marcu}{2000}]{marcu2000theory}
Marcu, D.
\newblock 2000.
\newblock The theory and practice of discourse parsing and summarisation.

\bibitem[\protect\citeauthoryear{Mikolov \bgroup et al\mbox.\egroup
  }{2014}]{mikolov2014word2vec}
Mikolov, T.; Chen, K.; Corrado, G.; and Dean, J.
\newblock 2014.
\newblock word2vec.

\bibitem[\protect\citeauthoryear{Mizuta and
  Collier}{2004}]{mizuta2004annotation}
Mizuta, Y., and Collier, N.
\newblock 2004.
\newblock An annotation scheme for a rhetorical analysis of biology articles.
\newblock In {\em LREC},  1737--1740.

\bibitem[\protect\citeauthoryear{Nawaz, Thompson, and
  Ananiadou}{2010}]{nawaz2010evaluating}
Nawaz, R.; Thompson, P.; and Ananiadou, S.
\newblock 2010.
\newblock Evaluating a meta-knowledge annotation scheme for bio-events.
\newblock In {\em Proceedings of the Workshop on Negation and Speculation in
  Natural Language Processing},  69--77.
\newblock Association for Computational Linguistics.

\bibitem[\protect\citeauthoryear{Nwogu}{1997}]{nwogu1997medical}
Nwogu, K.~N.
\newblock 1997.
\newblock The medical research paper: Structure and functions.
\newblock {\em English for specific purposes} 16(2):119--138.

\bibitem[\protect\citeauthoryear{Pennington, Socher, and
  Manning}{2014}]{pennington2014glove}
Pennington, J.; Socher, R.; and Manning, C.~D.
\newblock 2014.
\newblock Glove: Global vectors for word representation.
\newblock In {\em EMNLP}, volume~14,  1532--43.

\bibitem[\protect\citeauthoryear{Pyysalo \bgroup et al\mbox.\egroup
  }{2013}]{pyysalo2013distributional}
Pyysalo, S.; Ginter, F.; Moen, H.; Salakoski, T.; and Ananiadou, S.
\newblock 2013.
\newblock Distributional semantics resources for biomedical text processing.
\newblock {\em Proceedings of Languages in Biology and Medicine}.

\bibitem[\protect\citeauthoryear{Socher \bgroup et al\mbox.\egroup
  }{2013}]{socher2013parsing}
Socher, R.; Bauer, J.; Manning, C.~D.; and Ng, A.~Y.
\newblock 2013.
\newblock Parsing with compositional vector grammars.
\newblock In {\em In Proceedings of the ACL conference}.
\newblock Citeseer.

\bibitem[\protect\citeauthoryear{Svoboda and
  Reenstra}{2002}]{svoboda_approaches_2002}
Svoboda, K. K.~H., and Reenstra, W.~R.
\newblock 2002.
\newblock Approaches to studying cellular signaling: a primer for
  morphologists.
\newblock {\em The Anatomical record} 269(2):123--139.

\bibitem[\protect\citeauthoryear{Teufel and Moens}{1999}]{teufeldiscourse}
Teufel, S., and Moens, M.
\newblock 1999.
\newblock Discourse-level argumentation in scientific articles: human and
  automatic annotation.
\newblock In {\em Towards Standards and Tools for Discourse Tagging},  84--93.

\bibitem[\protect\citeauthoryear{Teufel and
  Moens}{2002}]{teufel2002summarizing}
Teufel, S., and Moens, M.
\newblock 2002.
\newblock Summarizing scientific articles: experiments with relevance and
  rhetorical status.
\newblock {\em Computational linguistics} 28(4):409--445.

\bibitem[\protect\citeauthoryear{Teufel}{2000}]{simone_teufel_argumentative_2000}
Teufel, S.
\newblock 2000.
\newblock {\em Argumentative {Zoning}: {Information} {Extraction} from
  {Scientific} {Text}.}
\newblock Ph.{D}. {Dissertation}, School of Cognitive Science, University of
  Edinburgh, Edinburg.

\bibitem[\protect\citeauthoryear{Vinyals \bgroup et al\mbox.\egroup
  }{2015}]{vinyals2015grammar}
Vinyals, O.; Kaiser, {\L}.; Koo, T.; Petrov, S.; Sutskever, I.; and Hinton, G.
\newblock 2015.
\newblock Grammar as a foreign language.
\newblock In {\em Advances in Neural Information Processing Systems},
  2755--2763.

\bibitem[\protect\citeauthoryear{Voit}{2012}]{Voit:2012:FCS:2341399}
Voit, E.
\newblock 2012.
\newblock {\em A First Course in Systems Biology}.
\newblock Garland Science, 1st edition.

\bibitem[\protect\citeauthoryear{Xu \bgroup et al\mbox.\egroup
  }{2015}]{xu2015show}
Xu, K.; Ba, J.; Kiros, R.; Courville, A.; Salakhutdinov, R.; Zemel, R.; and
  Bengio, Y.
\newblock 2015.
\newblock Show, attend and tell: Neural image caption generation with visual
  attention.
\newblock {\em arXiv preprint arXiv:1502.03044}.

\end{thebibliography}
\bibliographystyle{aaai}

\end{document}